\pdfoutput=1
\documentclass[runningheads]{llncs}
\usepackage{amssymb}
\usepackage{enumitem}
\usepackage{url}
\usepackage{marvosym}
\usepackage[T1]{fontenc}
\usepackage{graphicx}

\usepackage{amsmath}
\usepackage{multirow}
\begin{document}
\title{SBoRA: Low-Rank Adaptation with Regional Weight Updates}
%
%
\author{Lai-Man Po\inst{1} \and
Yuyang Liu\inst{1} \and
Haoxuan Wu\inst{1} \and
Tianqi Zhang\inst{1} \and
Wing-Yin Yu\inst{2} \and
Zhuohan Wang\inst{1} \and
Zeyu Jiang\inst{1} \and
Kun Li\inst{1}}
\authorrunning{LM Po et al.}
%
\institute{Department of Electrical Engineering, City University of Hong Kong, Kowloon, Hong Kong SAR, China \\
 \and
Huawei Noah's Ark Lab, Sha Tin, Hong Kong SAR, China \\
}
%
\maketitle      
\begin{abstract}

This paper introduces Standard Basis LoRA (SBoRA), a novel parameter-efficient fine-tuning approach for Large Language Models that builds upon the pioneering works of Low-Rank Adaptation (LoRA) and Orthogonal Adaptation. SBoRA reduces the number of trainable parameters by half or doubles the rank with the similar number of trainable parameters as LoRA, while improving learning performance. By utilizing orthogonal standard basis vectors to initialize one of the low-rank matrices (either $\mathbf{A}$ or $\mathbf{B}$), SBoRA facilitates regional weight updates and memory-efficient fine-tuning. This results in two variants, SBoRA-FA and SBoRA-FB, where only one of the matrices is updated, leading to a sparse update matrix $\mathrm{\Delta} \mathbf{W}$ with predominantly zero rows or columns. Consequently, most of the fine-tuned model’s weights $(\mathbf{W}_0+\mathrm{\Delta} \mathbf{W})$ remain unchanged from the pre-trained weights, akin to the modular organization of the human brain, which efficiently adapts to new tasks. Our empirical results demonstrate the superiority of SBoRA-FA over LoRA in various fine-tuning tasks, including commonsense reasoning and arithmetic reasoning. Furthermore, we evaluate the effectiveness of QSBoRA on quantized LLaMA models of varying scales, highlighting its potential for efficient adaptation to new tasks. Code is available at https://github.com/cityuhkai/SBoRA 

\end{abstract}
\section{Introduction}
\begin{figure}
\centering
\includegraphics[width=0.85\textwidth]{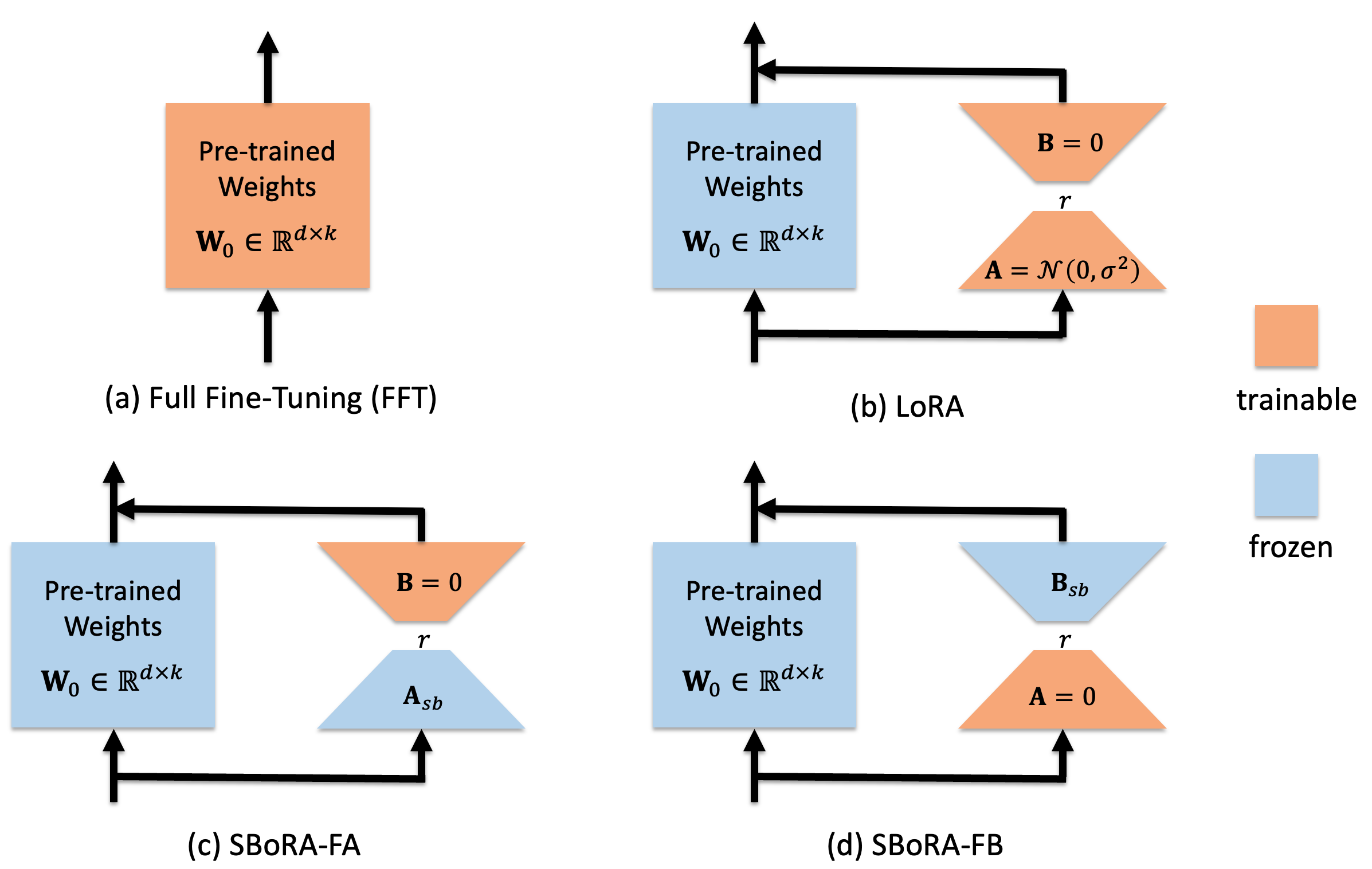}
\caption{Four fine-tuning strategies: (a) Full Fine-Tuning (FFT), (b) LoRA, (c) SBoRA-FA, and (d) SBoRA-FB.} \label{fig1}
\end{figure}

Large language models (LLMs) and diffusion models have become essential components of natural language processing \cite{brown2020language,touvron2023llama} and multimodal AI applications \cite{li2022blip,liu2024visual}. Full fine-tuning (FFT) has been shown to significantly improve their performance in various downstream tasks \cite{jin2024impact,wei2021finetuned} or introduce new concepts. However, fine-tuning pre-trained models, especially LLMs, by updating all parameters is computationally costly. As depicted in Fig.~\ref{fig1}(a), FFT involves directly updating the high-dimensional pre-trained weight matrix $\mathbf{W}_0$. 

To mitigate this challenge, parameter-efficient fine-tuning (PEFT) methods \cite{houlsby2019parameter} have garnered significant attention, focusing on updating only a small fraction of parameters, like adapter \cite{he2021towards,houlsby2019parameter,mahabadi2021parameter}, prompt tuning \cite{lester2021power,razdaibiedina2023residual,wang2023non}. Among these methods, Low-Rank Adaptation (LoRA) \cite{hu2021lora} has emerged as a groundbreaking technique for LLMs. LoRA introduces a parallel low-rank adapter to the weights of linear layers as shown in Fig.~\ref{fig1}(b), reducing memory overhead and computational costs during fine-tuning. However, there are still limitations to LoRA, including activation memory consumption and a performance gap compared to FFT. These limitations have led to the development of many LoRA variants \cite{chavan2023one,chen2023longlora,dettmers2024qlora,guo2023lq,liu2024dora,renduchintala2023tied,sheng2023s,wang2023multilora,xu2023qa,yuan2024mora,zhang2023lora}.

\vspace{-0.5cm}
\vspace{-0.5cm}


In this work, we introduce Standard Basis LoRA (SBoRA), a novel approach that further reduces the computational and memory requirements while enhancing the learning capacity. SBoRA achieves this by selectively updating specific rows or columns of the pre-trained weight matrix $\mathbf{W}_0$, preserving most of the base model's weights. The resulting update matrix $\mathrm{\Delta} \mathbf{W} = \mathbf{BA}$ mainly consists of zero rows or columns, indicating that most of the fine-tuned model's weights $(\mathbf{W}' = \mathbf{W}_0 + \mathrm{\Delta} \mathbf{W})$ remain identical to the pre-trained weights. This localized learning process mirrors the modular organization of the human brain, where specific cognitive functions are localized in distinct regions \cite{genon2018characterize}. For example, the hippocampus is responsible for episodic memory \cite{yonelinas2024role}. This design potentially improves knowledge retention and adaptation efficiency.


Let's assume that the pre-trained weight matrix $\mathbf{W}_0$, the projection-down matrix $\mathbf{A}$, and the projection-up matrix $\mathbf{B}$ are given. SBoRA adopts a unique approach, utilizing orthogonal standard basis vectors (one-hot vectors) to construct projection matrices $\mathbf{A}_{sb}$ (for SBoRA-FA) or $\mathbf{B}_{sb}$ for (SBoRA-FB), as shown in Figures~\ref{fig1}(c) and~\ref{fig1}(d). This approach eliminates the need to store coefficients for one projection matrix, reducing memory usage by approximately 50\%. In SBoRA-FA, only matrix $\mathbf{B}$ is updated during fine-tuning, with $\mathbf{A}_{sb}$ and $\mathbf{W}_0$ remaining frozen. The weight changes ($\mathrm{\Delta}\mathbf{W}=\mathbf{BA}_{sb}$) reside in a low-rank subspace spanned by $\mathbf{A}_{sb}$, and the projection-down transformation can be efficiently achieved by selecting a subset of input vector samples. SBoRA-FB operates similarly, updating only matrix $\mathbf{A}$ with a pre-defined $\mathbf{B}_{sb}$. Both variants align with LoRA, reducing computational overhead during fine-tuning without introducing extra inference latency.

We extensively validated SBoRA across diverse tasks, including evaluations on floating-point 16-bit precision models and quantized models in the QLoRA framework \cite{dettmers2024qlora}. For 16-bit models, we focused on commonsense reasoning and arithmetic reasoning tasks. Quantized models were evaluated using the MMLU benchmarks \cite{hendrycks2020measuring}. Experimental results consistently demonstrate SBoRA's superiority over other PEFT baselines. For instance, with similar number of trainable parameters (SBoRA: rank 64, LoRA/DoRA: rank 32), SBoRA-FA demonstrates significant improvements in commonsense reasoning (+2.9\%/1.7\% on LLaMA-7B/LLaMA3-8B) compared to LoRA, and arithmetic reasoning (+2.8\%/+2.0\% on LLaMA-7B/LLaMA3-8B) compared to DoRA. Furthermore, with approximately half the trainable parameters (QSBoRA/QLoRA/QDoRA: rank 64), QSBoRA-FA exhitbs notable enhancements on MMLU benchmarks, such as +4.3\%/+6.9\% on quantized LLaMA-13B/LLaMA3-8B compared to QLoRA.

The primary contributions of this work can be summarized as follows:
\begin{itemize}
    \item Introduces SBoRA as a novel PEFT approach for LLMs, which reduces the number of trainable parameters or doubles the rank while improving performance.
    \item Presents SBoRA-FA and SBoRA-FB, which use standard basis vectors for sparse updates, keeping most weights unchanged from the pre-trained model.
    \item Demonstrates that SBoRA-FA outperforms methods like LoRA and OA in reasoning tasks with half the trainable parameters, and matches DoRA's performance.
    \item Demonstrates effectiveness with quantized LLaMA models, showcasing versatility across different model types.
\end{itemize}

\section{Related Work}
\subsection{Full Fine-Tuning (FFT) and Parameter-Efficient Fine-Tuning (PEFT)}
Fine-tuning is a fundamental concept in deep learning that enables leveraging pre-trained models for new tasks. However, FFT has limitations, including high storage and computational costs. To address these limitations, PEFT techniques \cite{houlsby2019parameter} have been proposed, which adapt only a small subset of a model's parameters to a new task. Early approaches include the use of adapter layers, which add small, trainable modules to each Transformer layer. Another approach is prefix tuning \cite{liu2024visual}, which optimizes a few continuous "prefix" vectors prepended to the input sequence. Both of these approaches significantly reduce the memory and computation needed to fine-tune large models. However, they also have their limitations, such as extra inference latency in the case of adapter layers, and difficulty in optimization for prefix tuning.

\subsection{Low-Rank Adaptation(LoRA)}
LoRA \cite{hu2021lora} is a prominent PEFT method that approximates full-rank updates with low-rank updates within the FFT domain. Given a pre-trained parameter matrix $\mathbf{W}_0 \in \mathbb{R}^{d \times r}$, LoRA uses two low-rank matrices, $\mathbf{A} \in \mathbb{R}^{r \times k}$ and $\mathbf{B} \in \mathbb{R}^{d \times r}$, to compute the weight update $\mathrm{\Delta}\mathbf{W} \in \mathbb{R}^{d \times k}$. The output vector $\mathbf{h} \in \mathbb{R}^{d \times 1}$ of a LoRA linear layer can be expressed as:

\begin{equation}
    \mathbf{h} = \mathbf{W}_0\mathbf{x} + \mathrm{\Delta}\mathbf{Wx} = \mathbf{W}_0\mathbf{x} + \mathbf{BAx}  
\end{equation}

LoRA ensures $\mathrm{\Delta}\mathbf{W}$ is initialized to zero by initializing $\mathbf{A}$ with a uniform distribution and $\mathbf{B}$ with zero. The low-rank decomposition of $\mathrm{\Delta}\mathbf{W}$ into $\mathbf{BA}$ reduces the rank of $\mathrm{\Delta}\mathbf{W}$, making it more efficient than full-rank updating in FFT. LoRA's low-rank updating approach has demonstrated comparable performance to full-rank updating in tasks like text classification and instruction tuning, while reducing memory usage and computational requirements. It also simplifies deployment in multi-task scenarios and can match or exceed FFT performance using only a fraction of the total parameters.

\subsection{Variants of LoRA}
LoRA has sparked significant research in PEFT methods, including its own variants such as QLoRA \cite{dettmers2024qlora}, QA-LoRA \cite{xu2023qa}, LongLoRA \cite{chen2023longlora}, S-LoRA \cite{sheng2023s}, LQ-LoRA \cite{guo2023lq}, MultiLoRA \cite{wang2023multilora}, LoRA-FA \cite{zhang2023lora}, Tied-LoRA \cite{renduchintala2023tied}, GLoRA \cite{chavan2023one}, and DoRA \cite{liu2024dora}. QLoRA is an industry-standard technique for PEFT of LLMs, it employs 4-bit quantization on pretrained weights and trains LoRA modules on this quantized representation. Techniques like 4-Bit NormalFloat (NF4) Format, Double Quantization, and Paged Optimizers further minimize memory usage. QA-LoRA reduces the computational burden with group-wise quantization. LongLoRA enables fine-tuning for longer context lengths using sparse local attention. S-LoRA presents a scalable strategy for deploying multiple LoRA modules efficiently. LQ-LoRA refines the quantization scheme for improved performance. MultiLoRA handles complex multi-task learning, LoRA-FA reduces memory overhead, and Tied-LoRA leverages weight tying. GLoRA adapts both weights and activations, and DoRA decomposes weights for enhanced learning capacity. Recently, Orthogonal Adaptation enables efficient merging of customized models without sacrificing fidelity or incurring additional computational costs.


\section{Standard Basis Low-Rank Adaptation (SBoRA)}
SBoRA, inspired by Orthogonal Adaptation \cite{po2024orthogonal}, utilizes a predefined orthogonal basis $\mathbf{O}_A$ to generate orthogonal subspaces. These subspaces, represented as low-rank projection-down matrices $\mathbf{A}_\mathit{i}$, enable independent LoRA fine-tuning for multi-concept customization in diffusion models. The linear layer combining $c$ custom concepts, given a pre-trained weight $\mathbf{W}_0$, can be expressed as

\begin{equation}
    Linear(\mathbf{x}) = (\mathbf{W}_0 + \sum_{i=1}^c \lambda_i \cdot \mathbf{B}_i \mathbf{A}_i ) \mathbf{x}
\end{equation}

\noindent where $i$ denotes the index of the $i$-th LoRA, and $\lambda_i$ are scalar factors determined through empirical tuning. The input is a column vector $\mathbf{x} \in \mathbb{R}^{k \times 1}$. $\mathbf{A}_i \in \mathbb{R}^{r \times k}$ matrices are constructed by selecting $r$ non-overlapping orthogonal basis vectors from the shared orthogonal basis $\mathbf{O}_\mathit{A} \in \mathbb{R}^{k \times k}$. $\mathbf{O}_\mathit{A}$ consists of $k$ orthogonal basis vector $\mathbf{o}_p \in \mathbb{R}^{1 \times k}$, where $p=1,2,\ldots,k$. It can be represented as

\begin{equation}
    \mathbf{O}_\mathit{A} = 
    \left[
    \begin{array}{c}
          \mathbf{o}_1 \\ \mathbf{o}_2 \\ \vdots \\ \mathbf{o}_k
    \end{array}
    \right] 
    \mathrm{\ with\ } \mathbf{o}_p \mathbf{o}_q^T = 0 \mathrm{\ if \ } p \neq q
\end{equation}

To maintain orthogonality between projection-down matrices, these basis vectors must be non-overlapping from $\mathbf{O}_A$, ensuring $\mathbf{A}_p \mathbf{A}_q^T=0$ if $p \neq q$. And to minimize crosstalk, $\mathbf{B}_j \mathbf{A}_j \mathbf{x} = \mathbf{0}$ if $j \neq i$ when $\mathbf{x}$ belongs to subspace $\mathbf{A}_i$. During fine-tuning, the orthogonal matrices $\mathbf{A}_i$ are frozen while updating the projection-up matrices $\mathbf{B}_i$ to learn different concepts.

\subsection{Orthogonal Standard Basis}
The orthogonal adaptation approach offers not only effective multi-concept merging but also memory efficiency. During LoRA fine-tuning, the $\mathbf{A}_i$ matrices are frozen, which significantly reduces memory requirements for gradient and intermediate activation storage. This is similar to the LoRA-FA \cite{zhang2023lora}, where projection-down matrix $\mathbf{A}$ is randomly initialized and then frozen during training. However, we can further improve this orthogonal adaptation idea by utilizing a standard basis with one-hot vectors as basis vectors. In this case, the orthogonal basis $\mathbf{O}_A$ becomes a $k\times k$ identity matrix $\mathbf{I}$. Specifically, $\mathbf{O}_A$ can be represented as:

\begin{equation}
    \mathbf{O}_A = \mathbf{I} = 
    \left[
    \begin{array}{c}
         \mathbf{e}_1 \\
         \mathbf{e}_2 \\
         \vdots \\
         \mathbf{e}_k
    \end{array}
    \right]
    =
    \left[
    \begin{array}{cccc}
       1 & 0 & \cdots & 0 \\
       0 & 1 & \cdots & 0 \\
       \vdots & \vdots & \ddots & \vdots \\
       0 & 0 & 0 & 1
    \end{array}
    \right]
\end{equation}

\noindent where $\mathbf{e}_p\mathbf{e}_q^T = 0$ if $p \neq q$. The standard basis vector $\mathbf{e}_k \in \mathbb{R}^{1 \times k}$ are one-hot row vectors, each having a single non-zero entry of 1 at index $p$. For instance, $\mathbf{e}_p$ can be represented as

\vspace{-0.2cm}
\begin{equation}
\begin{aligned}
    \mathbf{e}_p = 
    [0 \ \cdots \ 0 &\ 1 \ 0 \ \cdots \ 0] \\[-3pt]
    &\uparrow \\[-5pt]
    &\ p
\end{aligned}
\end{equation}

\noindent where the non-zero component with value one at the index of $p$. In SBoRA-FA, we don’t need to randomly pre-generate an orthogonal-basis matrix $\mathbf{O}_A$. Instead, we randomly select $r$ non-overlapping indices between 1 and $k$ to construct a standard subspace basis matrix $\mathbf{A}_{sb}$ with $r$ standard basis row vectors of $\mathbf{e}_p$. This approach eliminates the need to store weights for the projection-down matrix, as only the indices of the standard basis vectors are required to represent $\mathbf{A}_{sb}$. The basic structure of SBoRA-FA is depicted in Fig.~\ref{fig1}(c), where both $\mathbf{W}_0$ and  $\mathbf{A}_{sb}$ are frozen, and only the matrix $\mathbf{B}$ is updated with initial values of zero that ensure that the pre-trained models do not alter the model prediction before fine-tuning. Similarity, SBoRA-FB freezes the projection-up as standard-basis low-rank matrix $\mathbf{B}_{sb}$ and only update the projection-down matrix $\mathbf{A}$ during the fine-tuning, in which orthogonal basis matrix $\mathbf{O}_B$ is a $d\times d$ identity matrix $\mathbf{I}$, the standard basis vectors are one-hot column vectors instead of row vectors. 

\begin{equation}
    \mathbf{O}_B = \mathbf{I} = 
    \left[
    \begin{array}{cccc}
         \mathbf{e}_1 &
         \mathbf{e}_2 &
         \cdots &
         \mathbf{e}_d
    \end{array}
    \right]
    =
    \left[
    \begin{array}{cccc}
       1 & 0 & \cdots & 0 \\
       0 & 1 & \cdots & 0 \\
       \vdots & \vdots & \ddots & \vdots \\
       0 & 0 & 0 & 1
    \end{array}
    \right]
\end{equation}

\noindent where $\mathbf{e}_p^T\mathbf{e}_q = 0$ if $p \neq q$. The standard basis vector $\mathbf{e}_p \in \mathbb{R}^{d \times 1}$ are one-hot column vectors, each having a single non-zero entry of 1 at index $p$. For instance, $\mathbf{e}_p$ can be represented as

\vspace{-0.2cm}
\begin{equation}
\begin{aligned}
    \mathbf{e}_p = 
    [0 \ \cdots \ 0 &\ 1 \ 0 \ \cdots \ 0]^T \\[-3pt]
    &\uparrow \\[-5pt]
    &\ p
\end{aligned}
\end{equation}

The basic structure of SBoRA-FB is depicted in Fig.~\ref{fig1}(d), where both $\mathbf{W}_0$ and  $\mathbf{B}_{sb}$ are frozen, and only the matrix $\mathbf{A}$ is updated with initial values of zero that ensure that the pre-trained models do not alter the model prediction before fine-tuning. 

\begin{figure}
\centering
    \includegraphics[width=0.9\textwidth]{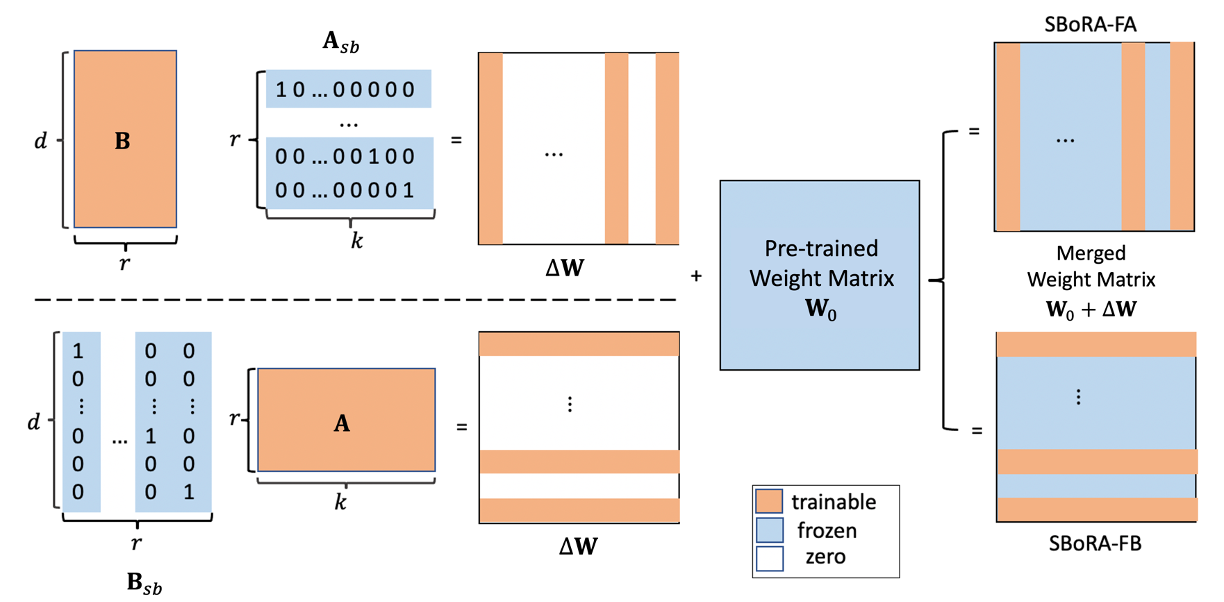}
    \caption{The diagram illustrates the regional weight update process of SBoRA, showcasing distinct $\mathbf{W}_0+\mathrm{\Delta}\mathbf{W}$ computing procedures of SBoRA-FA(upper) and SBoRA-FB (lower). The diagram employs different colors to represent frozen, trainable, and zero parameters.}
    \label{fig2}
\end{figure}

\subsection{Regional Weight Update}
An interesting property of the proposed SBoRA is that the merged weight matrix $\mathbf{W}'$ of the fine-tuned model is only regionally updated, with most of the weights remaining unchanged from the original pre-trained weight $\mathbf{W}_0$ as depicted in Fig.~\ref{fig2}. Specifically, the updated weight matrix $\mathbf{W}'$ of SBoRA-FA can be formally expressed as:

\vspace{-0.2cm}
\begin{equation}
    \mathbf{W}' = \mathbf{W}_0 + \mathrm{\Delta}\mathbf{W} = \mathbf{W}_0 + \mathbf{BA}_{sb}
\end{equation}

In which, the update matrix $\mathrm{\Delta}\mathbf{W}=\mathbf{BA}_{sb}$ is very sparse, with most of the columns having zero weights due to the one-hot nature of the standard basis subspace matrix $\mathbf{A}_{sb}$ as shown in the upper part of Fig.~\ref{fig2}. For example, when $r=2$ and $k=4$ with $\mathbf{A}_{sb}=[\mathbf{e}_1 \ \mathbf{e}_4]^T$, the $\mathrm{\Delta}\mathbf{W}$ will only have two non-zero column as

\begin{equation}
    \mathrm{\Delta}\mathbf{W}=\mathbf{BA}_{sb}=
    \left[
    \begin{array}{cc}
        b_{11} & b_{12} \\
        b_{21} & b_{22} \\
        b_{31} & b_{32} \\
        b_{41} & b_{42} 
    \end{array}
    \right]
    \left[
    \begin{array}{cccc}
        1 & 0 & 0 & 0 \\
        0 & 0 & 0 & 1 
    \end{array}
    \right]
    =
    \left[
    \begin{array}{cccc}
        b_{11} & 0 & 0 & b_{12} \\
        b_{21} & 0 & 0 & b_{22} \\
        b_{31} & 0 & 0 & b_{32} \\
        b_{41} & 0 & 0 & b_{42} 
    \end{array}
    \right]
\end{equation}

The fine-tuned weight $\mathbf{W}'$ is given by
\begin{equation}
    \mathbf{W}' = \mathbf{W}_0 + \mathrm{\Delta}\mathbf{W} =
    \left[
    \begin{array}{cccc}
        w_{11}+b_{11} & w_{12} & w_{13} & w_{14}+b_{12} \\
        w_{21}+b_{21} & w_{22} & w_{23} & w_{24}+b_{22} \\
        w_{31}+b_{31} & w_{32} & w_{33} & w_{34}+b_{32} \\
        w_{41}+b_{41} & w_{42} & w_{43} & w_{44}+b_{42} 
    \end{array}
    \right]
\end{equation}

For SBoRA-FB, when $r=2$ and $d=4$ with $\mathbf{B}_{sb} = [\mathbf{e}_1 \ \mathbf{e}_4]$, the $\mathrm{\Delta}\mathbf{W}$ will only have two non-zero row as

\begin{equation}
    \mathrm{\Delta}\mathbf{W}=\mathbf{B}_{sb}\mathbf{A}=
    \left[
    \begin{array}{cc}
        1 & 0 \\
        0 & 0 \\
        0 & 0 \\
        0 & 1
    \end{array}
    \right]
    \left[
    \begin{array}{cccc}
        b_{11} & b_{12} & b_{13} & b_{14} \\
        b_{12} & b_{22} & b_{23} & b_{24}
    \end{array}
    \right]
    =
    \left[
    \begin{array}{cccc}
        b_{11} & b_{12} & b_{13} & b_{14} \\
        0 & 0 & 0 & 0 \\
        0 & 0 & 0 & 0 \\
        b_{21} & b_{22} & b_{23} & b_{24}
    \end{array}
    \right]
\end{equation}

\vspace{0.25cm}
The fine-tuned weight $\mathbf{W}'$ is given by
\begin{equation}
    \mathbf{W}' = \mathbf{W}_0 + \mathrm{\Delta}\mathbf{W} =
    \left[
    \begin{array}{cccc}
        w_{11}+b_{11} & w_{12}+b_{12} & w_{13}+b_{13} & w_{14}+b_{14} \\
        w_{21} & w_{22} & w_{23} & w_{24} \\
        w_{31} & w_{32} & w_{33} & w_{34} \\
        w_{41}+b_{21} & w_{42}+b_{22} & w_{43}+b_{23} & w_{44}+b_{24}
    \end{array}
    \right]
\end{equation}

In these examples, only half of the weight matrix $\mathbf{W}'$ updated, specifically two columns or two rows $(r=2)$, while the remaining weights remain unchanged. In practice, $ r \ll $ min$(k,d)$, therefore only a small portion of the original weights will be updated. More details of the regional weight update property of SBoRA-FA and SBoRA-FB can be visualized in Fig.~\ref{fig2}. It resembles the localized learning process observed in neuroscience, where specific cognitive functions are associated with distinct brain regions. SBoRA efficiently adapts to new tasks by updating specific rows of the weight matrix, preserving existing knowledge, similar to how the brain reorganizes and refines neural connections in response to new experiences.

\section{Complexity Analysis of SBoRA}
\vspace{-0.1cm}


SBoRA can be implemented flexibility, with a proposed sampling-based multiplication (sampleMul) for improved computation and memory efficiency. While SBoRA can be implemented similarly to LoRA using matrix multiplication, it offers optimizations by representing the standard basis matrix $\mathbf{A}_{sb}$/$\mathbf{B}_{sb}$ as a $r$-dimensional vector with integers corresponding to the indices of the selected subspace basis vectors. These indices are initialized by randomly selecting $r$ unique numbers from 1 to $k$ for SBoRA-FA or from 1 to $d$ for SBoRA-FB. This implementation achieves approximately 50\% reduction in parameter storage and around half of the gradient storage due to the frozen standard basis matrix. Table~\ref{table1} compares the memory requirements of SBoRA and LoRA.
 
\vspace{-0.3cm}
\begin{table}
\caption{
Analysis of extra memory storage bring by a single LoRA module and SBoRA-FA/FB. Through the utilization of a standard basis and index-based updates SBoRA enables the replacement of an entire matrix with a 1-dimensional vector. This reduction leads to a significant decrease of approximately 50\% total parameters.
}\label{table1}
\begin{center}
\begin{tabular}{c|c|c|c}
\hline
{\bfseries Method} & {\bfseries Trainable Params} & {\bfseries Total Params} & {\bfseries Gradient Memory} \\
\hline
LoRA            & $(k+d) \times r$ & $(k + d) \times r$ & $(k+d) \times r$ \\
\hline
SBoRA-FA & $d \times r$ & $r+k \times r$ & $d \times r$   \\
\hline
SBoRA-FB & $k \times r$ & $r+d \times r$ & $k \times r$      \\
\hline
\end{tabular}
\end{center}
\end{table}
\vspace{-0.5cm}


SBoRA offers potential computation speedups through efficient implementation of matrix operations which leverages one-hot standard basis vectors. For SBoRA-FA, the multiplication $\mathbf{A}_{sb}\mathbf{x}$ can be efficiently implemented as a sampling operation ($\mathbf{x}[seq,\mathbf{A}_{sb}]$), which selects specific elements from $\mathbf{x}$ based on the indices in $\mathbf{A}_{sb}$. The forward process of SBoRA-FA with sampleMul can be expressed as:

\vspace{-0.5cm}
\begin{align}
    & \text{SBoRA-FA: } 
    \begin{cases}
    \mathbf{B}(\mathbf{A}_{sb}\mathbf{x})=\mathbf{B}(\mathbf{x}[seq, \mathbf{A}_{sb}]) \\
    \mathbf{h} = \mathbf{W}_0\mathbf{x}+\mathbf{BA}_{sb}\mathbf{x}
    \end{cases} \label{test1} \\
    \intertext{This sampling based multiplication (sampleMul) can be more efficient than standard matrix multiplication, especially for larger rank values. Similarly, for SBoRA-FB, the projection-up transformation can be implemented as a scatter operation using $index\_add$($\cdot$). The forward process of SBoRA-FB with sampleMul can be expressed as:}
    & \text{SBoRA-FB: } 
    \begin{cases}
    \mathbf{z}=\mathbf{Ax} \\
    \mathbf{h}=(\mathbf{W}_0\mathbf{x})\text{.}\textit{index\_add}(\mathbf{B}_{sb}, \mathbf{z})
    \end{cases} \label{test2}
\end{align}



As shown in the equation, projection-down vector $\mathbf{z}$ is directly added into the pre-trained weight optput through index addition method. 

These optimizations reduce computation complexity by saving one matrix multiplication operation compared to LoRA. During this process, two matrix multiplication operations $\mathbf{B}(\mathbf{A}_{sb}\mathbf{x})$ / $\mathbf{Ax}$ and $\mathbf{W}_0\mathbf{x}$, and one matrix addition operation of $\mathbf{W}_0\mathbf{x}+\mathbf{BA}_{sb}\mathbf{x}$ / $\mathbf{W}_0\mathbf{x}+\mathbf{B}_{sb}\mathbf{Ax}$ are involved. Table~\ref{table2} provides an analysis of computational requirements for SBoRA using sampleMul compared to LoRA, assuming input $\mathbf{x}$ in the shape of $[d\times 1]$.

\vspace{-0.3cm}
\begin{table}
\caption{Computational analysis and comparison of LoRA and SBoRA’s sampleMul implementation, with a pre-trained weight shape of $k\times d$ and a LoRA rank of $r$.}\label{table2}
\begin{center}
\begin{tabular}{c|c|c}
\hline
{\bfseries Method} & {\bfseries Multiplication} & {\bfseries Addition}  \\
\hline
\shortstack{LoRA}    & \shortstack{\ $k\times d$; $(k+d)\times r$ \ } & {\shortstack{\ $k \times (d-1)$; $r \times (d-1) + k \times (r-1)$; $1 \times k$}} \\
\hline
\shortstack{SBoRA-FA} & \shortstack{$k \times d$; $k \times r$} & \shortstack{$k \times (d-1)$; $k \times (r-1)$; $1\times k$} \\
\hline
\shortstack{SBoRA-FB} & \shortstack{$k \times d$; $d \times r$} & \shortstack{$k \times (d-1)$; $r \times (d-1)$; $1\times r$} \\
\hline
\end{tabular}
\end{center}
\end{table}
\vspace{-0.8cm}


\section{Experiment}
We conducted a comprehensive evaluation of SBoRA's effectiveness across a range of tasks. To begin with, we conducted a comparative analysis by fine-tuning LLaMA models \cite{touvron2023llama} of different versions and scales on both commonsense reasoning tasks and arithmetic reasoning tasks to assess the performance of SBoRA-FA and SBoRA-FB in comparison to other established PEFT methods. Furthermore, recognizing the growing interest in quantized models, we also evaluated SBoRA on the quantized versions of LLaMA models, utilizing the techniques introduced in QLoRA  \cite{dettmers2024qlora}.


\subsection{Evaluating SBoRA on Commonsense Reasoning Tasks}
We evaluated SBoRA-FA/FB against LoRA, DoRA, Orthogonal Adaptation (OA), and LoRA-FA approach on the LLaMA-7B/LLaMA3-8B language models for commonsense reasoning tasks. We followed the evaluation method introduced in DoRA \cite{liu2024dora} and LLM-Adapter \cite{hu2023llm} which used a comprehensive training dataset created by aggregating eight tasks, evaluating models on individual testing datasets for each task. The results of GPT-3.5 are included for reference as shown in Table~\ref{table3}.

For a comprehensive comparison, we initially fine-tuned models with SBoRA-FA/FB, OA, and LoRA-FA, both using a double rank of 64 to maintain a nearly identical number of trainable parameters (TP) as LoRA and DoRA (rank=32), with only additional storage for standard basis indices. We then conducted experiments with rank 32 for SBoRA-FA/FB, OA, and LoRA-FA, and rank 64 for other baselines to investigate if SBoRA and OA could achieve better performance with half the trainable parameters. Consistency was maintained by using one training epoch and learning rate of 2e-4 for all experiments.

\vspace{-0.3cm}
\begin{table}
\caption{
Comparative analysis of LLaMA-7B/LLaMA3-8B with different Parameter-Efficient Fine-Tuning methods, evaluating on the commonsense reasoning task. Results of GPT-3.5 are provided in the first row for reference. We report accuracy (\%) for eight sub-tasks and average accuracy (\%), with higher values indicating better performance. Column headers denote TP for trainable parameters and $r$ for rank.
}\label{table3}
\centering
   \scriptsize
   \begin{tabular}{c|c|c|c|c c c c c c c c c}
   \hline
    {Model} & {Method} & {$r$} & {TP} & {BoolQ} & {PIQA} & {SIQA} & {HS} & {WG} & {ARC-e} & {ARC-c} & {OBQA} &{Avg} \\
    \hline
    GPT-3.5 & - & - & - & 73.1 & 85.4 & 68.5 & 78.5 & 66.1 & 89.8 & 79.9 & 74.8 & 77.0 \\
    \hline
    \multirow{12}*{LLaMA-7B} & LoRA & \multirow{6}*{32} & 56.1M & 66.8 & 81.1 & 78.4 & 53.5 & 80.5 & 81.1 & 61.9 & 79.4 & \bfseries 72.8 \\
    ~ & DoRA & ~ & 57.0M & 68.8 & 82.0 & 70.6 & 57.6 & 73.2 & 79.4 & 64.2 & 78.2 & 71.7 \\
    ~ & OA & ~ & 28.0M & 65.4 & 80.5 & 76.8 & 46.2 & 78.9 & 79.4 & 61.7 & 75.4 & 70.5 \\
    ~ & LoRA-FA & ~ & 28.0M & 60.1 & 56.5 & 74.5 & 40.1 & 74.7 & 80.0 & 58.1 & 72.6 & 64.6 \\
    ~ & SBoRA-FA & ~ & 28.0M & 68.0 & 79.7 & 76.2 & 54.4 & 79.1 & 79.8 & 61.3 & 75.0 & 71.7 \\
    ~ & SBoRA-FB & ~ & 28.0M & 66.1 & 64.2 & 74.8 & 57.2 & 71.5 & 80.4 & 62.4 & 75.8 & 64.9 \\
    \cline{2-13}
    ~ & LoRA & \multirow{6}*{64} & 112.2M & 62.1 & 81.8 & 78.2 & 62.9 & 78.6 & 79.8 & 63.7 & 81.2 & 73.5 \\
    ~ & DoRA & ~ & 113.1M & 68.7 & 82.8 & 78.2 & 64.8 & 62.9 & 79.7 & 64.8 & 80.0 & 72.7 \\
    ~ & OA & ~ & 56.1M & 68.8 & 81.6 & 77.4 & 36.5 & 78.4 & 80.8 & 64.8 & 77.6 & 70.7 \\
    ~ & LoRA-FA & ~ & 56.1M & 64.9 & 79.9 & 76.8 & 33.6 & 69.5 & 79.8 & 62.1 & 75.4 & 67.8 \\
    ~ & SBoRA-FA & ~ & 56.1M & 68.2 & 81.3 & 77.6 & 74.7 & 81.1 & 80.8 & 62.8 & 79.4 & \bfseries 75.7 \\
    ~ & SBoRA-FB & ~ & 56.1M & 66.5 & 79.2 & 76.7 & 59.2 & 76.5 & 76.8 & 59.0 & 74.4 & 71.0 \\
    \hline
    \multirow{12}*{LLaMA3-8B} & LoRA & \multirow{6}*{32} & 56.6M & 71.9 & 86.7 & 80.4 & 94.0 & 85.6 & 87.8 & 75.9 & 83.6 & 83.2 \\
    ~ & DoRA & ~ & 57.4M & 73.6 & 87.1 & 80.8 & 94.4 & 86.1 & 88.8 & 78.3 & 84.2 & 84.2 \\
    ~ & OA & ~ & 25.2M & 68.2 & 87.1 & 80.3 & 94.6 & 87.8 & 90.0 & 79.7 & 87 & 84.3 \\
    ~ & LoRA-FA & ~ & 25.2M & 72.5 & 88.1 & 80.2 & 93.9 & 84.5 & 90.4 & 78.4 & 84.8 & 84.1 \\ 
    ~ & SBoRA-FA & ~ & 25.2M & 73.3 & 87.8 & 79.1 & 93.9 & 85.2 & 89.9 & 80.0 & 86.0 & \bfseries 84.4 \\
    ~ & SBoRA-FB & ~ & 31.5M & 72.9 & 86.3 & 78.8 & 92.6 & 83.0 & 88.8 & 76.3 & 85.0 & 83.0 \\
    \cline{2-13}
    ~ & LoRA & \multirow{6}*{64} & 113.2M & 72.5 & 87.8 & 80.3 & 94.4 & 86.4 & 88.7 & 79.3 & 85.2 & 84.3 \\
    ~ & DoRA & ~ & 114.0M & 70.5 & 86.0 & 80.3 & 91.8 & 83.7 & 86.2 & 74.7 & 83.2 & 82.1 \\
    ~ & OA & ~ & 50.3M & 74.2 & 88.6 & 81.3 & 72.2 & 86.0 & 88.6 & 78.3 & 84.4 & 81.7 \\
    ~ & LoRA-FA & ~ & 50.3M & 73.9 & 86.7 & 80.0 & 94.6 & 86.7 & 88.9 & 78.4 & 83.8 & 84.1 \\
    ~ & SBoRA-FA & ~ & 50.3M & 74.0 & 88.3 & 80.8 & 94.3 & 86.3 & 89.9 & 78.7 & 86.6 & \bfseries 84.9 \\
    ~ & SBoRA-FB & ~ & 62.9M & 71.8 & 85.2 & 79.2 & 91.4 & 82.9 & 86.7 & 74.0 & 83.4 & 81.8 \\
    \hline
   \end{tabular} 
\end{table}
\vspace{-0.4cm}

In LLaMA-7B, SBoRA-FA (rank=64) outperforms all baseline methods with a similar number of trainable parameters, including LoRA (rank=32), DoRA (rank=32), OA (rank=64) and LoRA-FA (rank=64). SBoRA-FA with rank 32 demonstrates commendable performance by still outperforming OA (rank=32) and LoRA-FA (rank=32), while achieving the same accuracy as DoRA (rank=32). It lags behind the highest accuracy achieved by LoRA (rank=32) by only 1.1\%. Despite having approximately half the trainable parameters, SBoRA performs comparably to models with the same rank. In LLaMA3-8B, the top two results are achieved by SBoRA-FA with rank 64 and 32, respectively.

\subsection{Evaluating SBoRA on Arithmetic Reasoning}

We also evaluated the effectiveness of SBoRA on arithmetic reasoning tasks using the LLaMA-7B and LLaMA3-8B models. We constructed the fine-tuning and evaluation datasets following the dataset settings in LLM-Adapters \cite{hu2023llm}. We performed tests with ranks 32 and 64. Since the training set for arithmetic reasoning is smaller (10k) compared to the commonsense reasoning training set, we increased the number of training epochs to 3 for better convergence and set the learning rate to 3e-4.

Table~\ref{table4} presents the performance of PEFT methods, including GPT-3.5. On average, for LLaMA-7B, GPT-3.5 (175B) outperforms adapter-based PEFT LLMs in accuracy. However, for simpler math reasoning tasks like MultiArith, adapter-based methods outperform GPT-3.5, and SBoRA-FA achieves the best performance. Initially, SBoRA-FA/FB, OA, and LoRA-FA used a rank of 64 for consistency in trainable parameters. SBoRA-FA achieves the best average accuracy. Increasing the rank of other baselines to 64 improves accuracy but still lags behind SBoRA-FA. We then reduced the rank of SBoRA, OA, and LoRA-FA to 32, it can be found that SBoRA maintains comparable performance to rank 32 baselines. Notably, for the challenging task AQuA, SBoRA-FA with rank 32 excels. Moving to LLaMA3-8B, SBoRA-FA with rank 32 achieves the best performance, followed by SBoRA-FA with rank 64. Notably, SBoRA-FA consistently outperforms OA and LoRA-FA with both rank 32 and 64.

\begin{table}
\caption{
Comparative analysis of LLaMA-7B/LLaMA3-8B with different Parameter-Efficient Fine-Tuning methods, evaluating on the arithmetic reasoning tasks. Results of GPT-3.5 are provided in the first row for reference. We report accuracy (\%) for eight sub-tasks and average accuracy (\%), with higher values indicating better performance. Column headers denote TP for trainable parameters and $r$ for rank.
}
    \label{table4}
    \centering
    \scriptsize
    \begin{tabular}{c|c|c|c|c c c c c c c}
        \hline
        LLM & Method & Rank & TP & MultiArith & GSM8K & AddSub & AQuA & SingleEq & SVAMP & Avg \\
        \hline
        GPT-3.5 & - & - & - & 83.8 & 56.4 & 85.3 & 38.9 & 88.1 & 69.9 & 70.4 \\
        \hline
        \multirow{12}*{LLaMA-7B} & LoRA & \multirow{6}*{32} & 56.1M & 94.5 & 36.3 & 81.8 & 15.0 & 82.7 & 45.6 & \bfseries 59.3 \\
        ~ & DoRA & ~ & 57.0M & 95.7 & 36.2 & 78.7 & 15.4 & 81.7 & 46.6 & 59.1 \\
        ~ & OA & ~ & 28.0M & 96.2 & 37.5 & 76.7 & 15.0 & 77.4 & 41.7 & 57.4 \\
        ~ & LoRA-FA & ~ & 28.0M & 95.2 & 35.3 & 79.0 & 16.5 & 77.6 & 46.1 & 58.3 \\
        ~ & SBoRA-FA & ~ & 28.0M & 95.5 & 34.6 & 79.7 & 20.1 & 78.9 & 44.8 & 58.9 \\
        ~ & SBoRA-FB & ~ & 28.0M & 92.2 & 31.0 & 77.5 & 15.7 & 78.5 & 41.8 & 56.1 \\
        \cline{2-11}
        ~ & LoRA & \multirow{6}*{64} & 112.2M & 94.0 & 36.8 & 84.3 & 17.3 & 82.3 & 44.7 & 59.9 \\
        ~ & DoRA & ~ & 113.1M & 95.0 & 35.5 & 84.1 & 20.1 & 85.0 & 47.1 & 61.1 \\
        ~ & OA & ~ & 56.1M & 95.7 & 37.5 & 79.5 & 16.5 & 80.7 & 46.7 & 59.4 \\
        ~ & LoRA-FA & ~ & 56.1M & 94.7 & 35.6 & 80.2 & 17.3 & 80.9 & 49.9 & 59.8 \\
        ~ & SBoRA-FA & ~ & 56.1M & 97.8 & 36.6 & 85.1 & 19.3 & 83.9 & 48.5 & \bfseries 61.9 \\
        ~ & SBoRA-FB & ~ & 56.1M & 94.8 & 33.1 & 77.5 & 16.9 & 78.5 & 40.6 & 56.9 \\
        \hline 
        \multirow{12}*{LLaMA3-8B} & LoRA & \multirow{6}*{32} & 56.6M & 68.3 & 50.5 & 83.3 & 35.8 & 87.2 & 71.2 & 66.1 \\
        ~ & DoRA & ~ & 57.4M & 97.3 & 62.0 & 90.9 & 25.6 & 94.9 & 73.4 & 74.0 \\
        ~ & OA & ~ & 25.2M & 98.3 & 65.5 & 93.2 & 28.0 & 96.3 & 75.7 & 76.2 \\
        ~ & LoRA-FA & ~ & 25.2M & 71.1 & 86.3 & 58.1 & 67.7 & 86.0 & 29.5 & 66.5 \\
        ~ & SBoRA-FA & ~ & 25.2M & 99.5 & 66.0 & 91.9 & 30.3 & 97.4 & 75.8 & \bfseries 76.8 \\
        ~ & SBoRA-FB & ~ & 31.5M & 98.0 & 57.2 & 92.2 & 33.9 & 94.1 & 69.6 & 74.2 \\
        \cline{2-11}
        ~ & LoRA & \multirow{5}*{64} & 113.2M & 97.2 & 56.3 & 92.7 & 22.8 & 92.3 & 69.3 & 71.8 \\
        ~ & DoRA & ~ & 114.0M & 97.8 & 55.2 & 91.1 & 24.0 & 94.7 & 72.0 & 72.5 \\
        ~ & OA & ~ & 50.3M & 98.3 & 62.5 & 91.4 & 26.0 & 96.2 & 72.2 & 74.4 \\
        ~ & LoRA-FA & ~ & 50.3M & 70.3 & 52.9 & 84.1 & 28.0 & 87.0 & 70.3 & 65.4 \\
        ~ & SBoRA-FA & ~ & 50.3M & 99.2 & 64.7 & 94.4 & 24.8 & 98.0 & 75.0 & \bfseries 76.0 \\
        ~ & SBoRA-FB & ~ & 62.9M & 98.2 & 50.9 & 87.1 & 28.0 & 91.7 & 63.0 & 69.8 \\
        \hline
    \end{tabular}
\end{table}

SBoRA-FA/FB achieves comparable or superior results to other baseline PEFT methods with half or same the trainable parameters for the same rank. Arithmetic reasoning tasks are particularly well-suited for SBoRA due to the specialized knowledge required in mathematics. SBoRA's regional weight update property makes it more suitable for higher ranks like 64, impacting more pre-trained weight regions.

Based on experimental observations, SBoRA has been shown to significantly reduce training time compared to DoRA. To further illustrate the computational and memory efficiency advantages of SBoRA, we conducted a detailed analysis of the time cost and GPU memory usage during the training process for arithmetic reasoning tasks. The results are visually represented in the accompanying diagrams, which were performed on an NVIDIA RTX 4090.

\begin{figure}
\centering
    \includegraphics[width=0.95\textwidth]{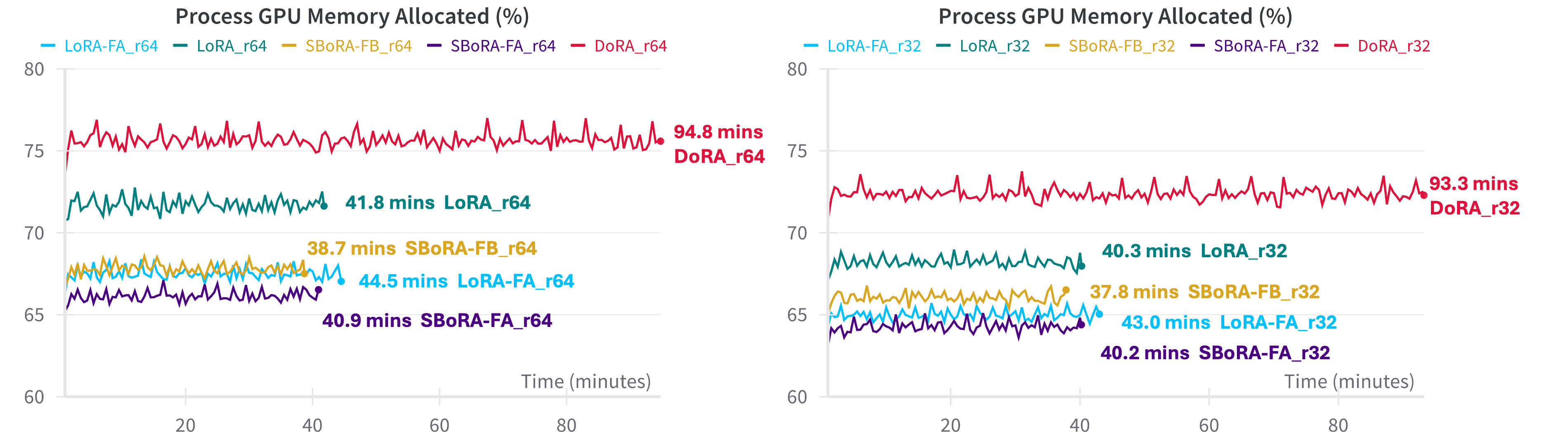}
    \caption{
    GPU usage and training time for LLaMA-7B on arithmetic reasoning tasks. Results for rank 64 (left) and 32 (right) are displayed. Y-axis: GPU usage; X-axis: training time. Total training time is labeled for each method.
    }
    \label{fig3}
\end{figure}

\begin{figure}
\centering
    \includegraphics[width=0.95\textwidth]{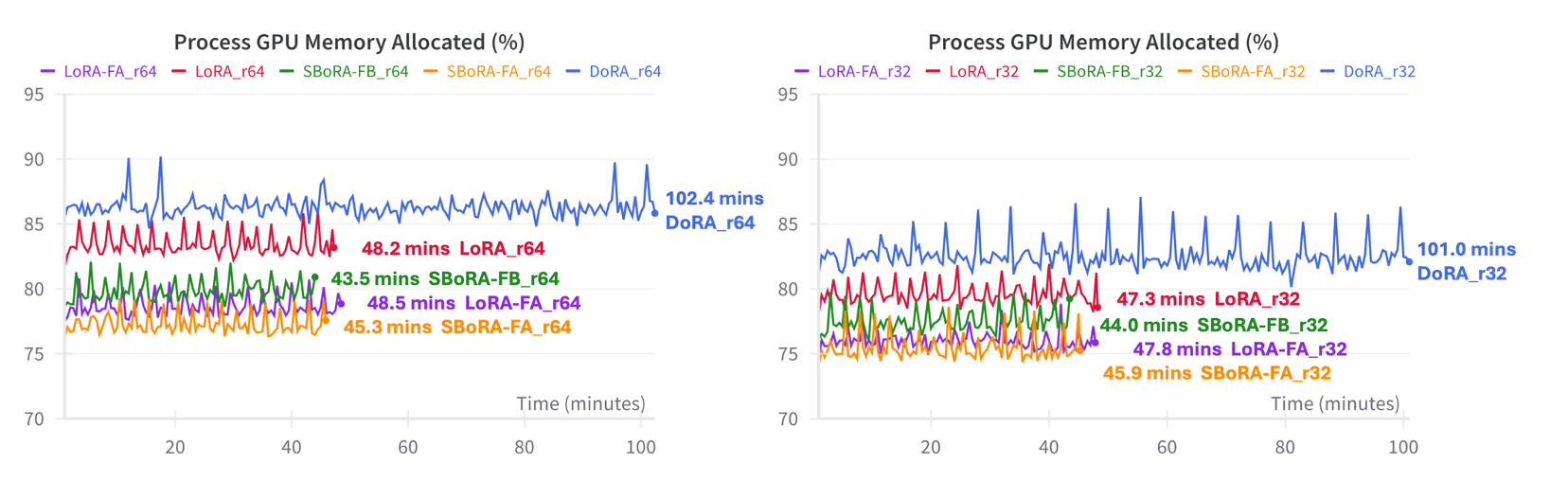}
    \caption{
    GPU usage and training time for LLaMA3-8B on arithmetic reasoning tasks. Results for rank 64 (left) and 32 (right) are displayed. Y-axis: GPU usage; X-axis: training time. Total training time is labeled for each method.
    }
    \label{fig4}
\end{figure}

We report the training time and GPU memory allocation for both LLaMA-7B and LLaMA3-8B models, comparing SBoRA with LoRA, LoRA-FA, and DoRA using ranks of 32 and 64 for each method. Figures~\ref{fig3} and~\ref{fig4} present the results for LLaMA-7B and LLaMA3-8B, respectively. The diagrams clearly demonstrate that, at equivalent ranks, SBoRA exhibits reduced training time and memory usage compared to both LoRA and DoRA during the training process. Notably, SBoRA-FA and SBoRA-FB achieve a significant reduction in training time, more than halving the duration required by DoRA.

\vspace{-0.4cm}

\subsection{QSBoRA Evaluation on MMLU}
In this section, we explored the effectiveness of SBoRAs in the memory-efficient QLoRA framework \cite{dettmers2024qlora}. QLoRA utilizes techniques such as 4-bit NormalFloat (NF4) double quantization, gradient checkpointing, and paged optimizer to reduce memory usage. Following the experiments in QLoRA, we fine-tuned the NF4 quantized versions of LLaMA-7B/13B and LLaMA3-8B using the Alpaca \cite{stanford_alpaca} and sampled FLAN v2 \cite{wei2021finetuned} datasets. We evaluated model performance on the MMLU benchmark \cite{hendrycks2020measuring} by comparing the mean 5-shot MMLU test accuracy.

\begin{table}
\caption{Finetune LLaMA-7B/13B and LLaMA3-8B on Alpaca and Flan v2. We measured the performance using MMLU benchmark and report the 5-shot average accuracy. The training sets and the number of trainable parameters (TP) are included.}
    \label{table5}
    \centering
    \scriptsize
    \begin{tabular}{c|c|c| c c|c| c c|c| c c}
    \hline
        \multicolumn{2}{c|}{NFloat4} & \multicolumn{3}{|c|}{LLaMA-7B} & \multicolumn{3}{|c|}{LLaMA-13B} & \multicolumn{3}{|c}{LLaMA3-8B} \\
        \hline
        PEFT & $r$ & TP & Alpaca & Flanv2 & TP & Alpaca & Flanv2 & TP & Alpaca & Flanv2 \\
        \hline
        QLoRA & \multirow{6}*{64} & 80.0M & 37.9 & 44.4 & 125.2M & 45.4 & 46.7 & 83.9M & 51.9 & 49.5 \\
        QDoRA & ~ & 80.6M & 38.0 & 42.8 & 126.2M & 46.7 & 48.8 & 84.6M & 53.0 & 51.9  \\
        QOA & ~ & 43.5M & 37.6 & 44.0 & 68.2M & 47.8 & 50.4 & 44.0M & 54.4 & 54.8 \\
        QLoRA-FA & ~ & 43.5M & 36.5 & 43.1 & 68.2M & 47.9 & 49.4 & 44.0M & 55.1 & 55.5 \\
        QSBoRA-FA & ~ & 43.5M & 36.5 & 43.1 & 68.2M & 49.0 & 51.0 & 44.0M & \bfseries 56.5 & \bfseries 56.4 \\
        QSBoRA-FB & ~ & 36.4M & 36.9 & 43.4 & 57.0M & 48.3 & 50.5 & 39.8M & 54.5 & 55.0 \\
        \hline
        QSBoRA-FA & \multirow{2}*{128} & 87.0M & \bfseries 39.3 & \bfseries 45.8 & 136.3M & \bfseries 50.0 & \bfseries51.7 & 88.1M & 56.3 & 54.2 \\
        QSBoRA-FB & ~ & 72.9M & 39.0 & 45.7 & 114.0M & 49.5 & 50.5 & 79.7M & 54.5 & 52.4 \\
        \hline
    \end{tabular}
\end{table}

To ensure fairness, we used the same hyperparameter settings for all experiments. For the Alpaca dataset, we used a batch size (bs) of 16 and 10k training steps. For FLAN v2, we used a batch size 8 and 20k training steps. In the QLoRA framework, we initially used a rank of 64 for all methods, including SBoRA-FA/FB, LoRA, DoRA, OA, and LoRA-FA. We then doubled the rank to 128 for SBoRA-FA/FB to maintain a similar number of trainable parameters as LoRA and DoRA with rank 64. We added adapters to linear layers in both self-attention and MLP modules.

The results in Table~\ref{table5} show that within LLaMA-7B, both QSBoRA-FA/FB with rank 64 achieve comparable accuracy results to QLoRA. In the case of LLaMA-13B and LLaMA3-8B, QSBoRA-FA with rank 64 outperforms other baseline methods in terms of accuracy, indicating its compatibility with larger-scale models. Furthermore, increasing the rank to 128 leads to improved performance for both QSBoRA-FA and QSBoRA-FB in LLaMA-7B and LLaMA-13B.

\section{Conclusion}

We presented SBoRA, an innovative PEFT method for LLMs that boosts learning capacity while cutting down on memory and computational needs. Our experiments demonstrate that SBoRA-FA excels in both commonsense and arithmetic reasoning tasks, even when using roughly half the number of trainable parameters, and significantly reduces training time and GPU memory usage. Empirical results indicate that SBoRA-FA surpasses methods like LoRA, LoRA-FA, and OA, and matches the performance of DoRA, making it an excellent choice for single-task fine-tuning. Within the QLoRA framework, SBoRA-FA is highly compatible with quantized models and further enhances performance at higher ranks, proving to be a robust, versatile, and resource-efficient fine-tuning method.

SBoRA efficiently adapts to new tasks while preserving pre-trained weights, akin to the modular organization of the brain. This approach could inspire AI architectures that emulate biological neural systems. Future work includes developing Multi-SBoRA, which enables independent fine-tuning for each task, minimizing interference and maximizing task-specific knowledge. This advancement could lead to more efficient AI systems with distinct capabilities across multiple tasks.

%
%
%
\bibliographystyle{splncs04}

\end{document}